\documentclass[10pt,twocolumn,letterpaper]{article}

\usepackage[pagenumbers]{cvpr} 
\usepackage[dvipsnames]{xcolor}

\usepackage{graphicx}
\usepackage{amsmath}
\usepackage{amssymb}
\usepackage{booktabs}
\usepackage{multirow}
\usepackage{color}
\usepackage{pifont}
\usepackage[T1]{fontenc}
\usepackage{caption}
\makeatletter

\newcommand{\Rmnum}[1]{\expandafter\@slowromancap\romannumeral #1@}
\makeatother

\definecolor{cvprblue}{rgb}{0.21,0.49,0.74}
\usepackage[pagebackref,breaklinks,colorlinks,citecolor=cvprblue]{hyperref}

\def\paperID{3282} 
\def\confName{CVPR}
\def\confYear{2024}

\title{BSNet: Box-Supervised Simulation-assisted Mean Teacher for 3D Instance Segmentation}

\author{Jiahao Lu$^{1}$, Jiacheng Deng$^1$, Tianzhu Zhang$^{1,2,}$\footnotemark[1]~ \\
\small{$^1$University of Science and Technology of China, $^2$Deep Space Exploration Lab}\\
{\tt\small \{lujiahao, dengjc\}@mail.ustc.edu.cn, {tzzhang@ustc.edu.cn}}\\
}

\begin{document}
\maketitle
\renewcommand{\thefootnote}{\fnsymbol{footnote}}
\footnotetext[1]{Corresponding Author}
\begin{abstract}
3D instance segmentation (3DIS) is a crucial task, but point-level annotations are tedious in fully supervised settings.
Thus, using bounding boxes (bboxes) as annotations has shown great potential. %
The current mainstream approach is a two-step process, involving the generation of pseudo-labels from box annotations and the training of a 3DIS network with the pseudo-labels.
However, due to the presence of intersections among bboxes, not every point has a determined instance label, especially in overlapping areas.
To generate higher quality pseudo-labels and achieve more precise weakly supervised 3DIS results, we propose the Box-Supervised Simulation-assisted Mean Teacher for 3D Instance
Segmentation (BSNet), which devises a novel pseudo-labeler called Simulation-assisted Transformer.
The labeler consists of two main components. 
The first is Simulation-assisted Mean Teacher, which introduces Mean Teacher for the first time in this task and constructs simulated samples to assist the labeler in acquiring prior knowledge about overlapping areas. 
To better model local-global structure, we also propose Local-Global Aware Attention as the decoder for teacher and student labelers.
Extensive experiments conducted on the ScanNetV2 and S3DIS datasets verify the superiority of our designs.  Code is available at \href{https://github.com/peoplelu/BSNet}{https://github.com/peoplelu/BSNet}.
 
\end{abstract}
\vspace{-1em}
\section{Introduction}
3D instance segmentation is a fundamental task in 3D scene understanding, primarily focused on predicting masks and categories for every foreground object within a scene. 
Current instance segmentation methods are mainly in fully supervised settings~\cite{vu2022softgroup,sun2023superpoint,ngo2023isbnet,schult2023mask3d,lu2023query} and achieve commendable results. 
However, the time-consuming nature of point-level annotations poses a significant challenge.
In contrast, annotating instances with 3D bboxes (object-level) is notably easier, requiring only the annotations for center points and dimensions (length, width, height). 
Nevertheless, a notable limitation stems from the use of bboxes, which cannot capture the detailed shape or geometry of objects. 
Consequently, bridging the gap between object-level and point-level annotations remains a challenge.
\begin{figure}[!t]
    \centering
    \includegraphics[width=1\linewidth]{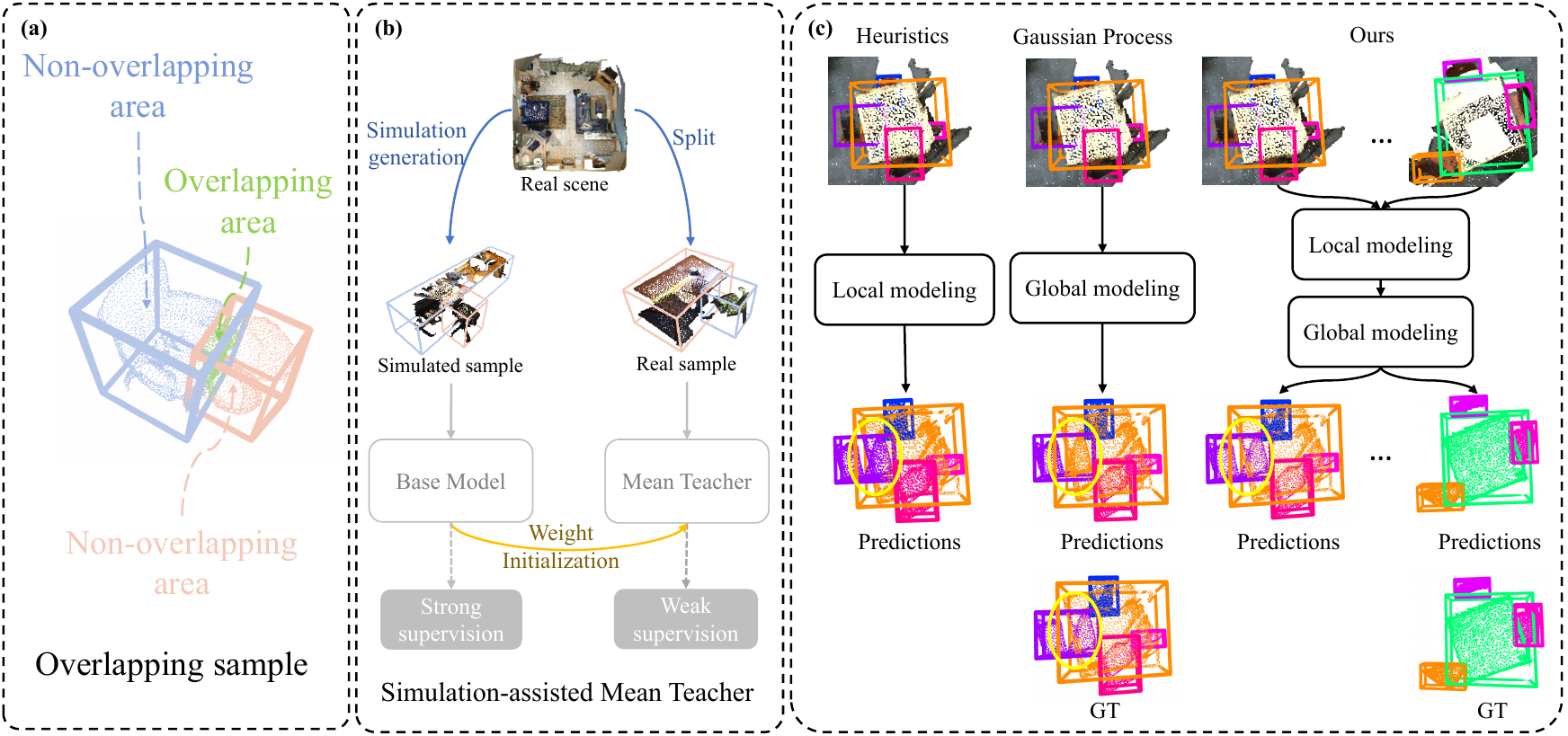}
    \vspace{-2em}
    \caption{
      (a) The visualization of an overlapping sample. (b) The proposed Simulation-assisted Mean Teacher helps the labeler acquire prior knowledge from simulated samples. (c) Our method improves local-global structure modeling of overlapping samples to generate better pseudo-labels (especially in yellow circles).}
    \label{fig:intro}
    \vspace{-1.6em}
\end{figure}

To solve the above challenge, existing methods~\cite{chibane2022box2mask,du2023weakly,shin2017real} conduct several explorations.
Box2Mask~\cite{chibane2022box2mask} parameterizes bboxes and utilizes them as labels.
However, due to bbox overlap, some point clouds may exist within multiple bboxes, introducing ambiguity in point-object assignments. 
As illustrated in Figure~\ref{fig:intro}(a), instance labels for point clouds in non-overlapping areas are determined as they only belong to one bbox. 
In contrast, overlapping areas are governed by two different bboxes, resulting in indeterminate instance labels.
Consequently, the point-wise predicted bboxes cannot be reliably used for clustering.
To better address the ambiguity in overlapping areas, WISGP~\cite{du2023weakly} employs straightforward heuristics based on local structure modeling. 
Concretely, for each indeterminate point, WISGP selects the most common label from its neighboring points as its label.
On the other hand, Gapro~\cite{ngo2023gapro} uses Gaussian Process (GP)~\cite{shin2017real} to train individual overlapping samples, modeling global structure by fitting the similarity relationships between all points into a Gaussian distribution.
Then Gapro computes posterior probability to achieve binary classification for overlapping areas.

Based on the preceding discussion, we identify two key issues in bbox-supervised 3DIS:
1) \textit{How to generate labels for overlapping areas?}
Using Mean Teacher~\cite{cao2023contrastive,li2022cross,zheng2021exploiting,deng2023se} to generate and continuously optimize pseudo-labels for overlapping areas is an effective way.
This paradigm allows for the online update of pseudo-labels during the training process, continuously transferring knowledge from a teacher network to a student network. 
Besides, the teacher network employs Exponential Moving Average (EMA) to integrate information from historical students, providing more stable learning targets for the student network.
However, given that non-overlapping areas contain a single object with a clear structure, while overlapping areas involve two intertwined objects, there is a significant disparity in complexity between them.
Hence, it is difficult to infer accurate pseudo-labels for overlapping areas solely according to non-overlapping area labels. 
To tackle this issue, given the abundance of non-overlapping bboxes in the dataset with definite labels, we can construct simulated overlapping samples using these bboxes. 
As illustrated in Figure~\ref{fig:intro}(b), we can train a base model on these simulated samples, and transfer this model to real datasets.
By this way, the information loss resulting from the absence of labels in overlapping areas can be compensated and higher quality pseudo-labels can be predicted.
2) \textit{How to better model structure of overlapping samples?}
As illustrated in Figure~\ref{fig:intro}(c), current methods~\cite{du2023weakly,ngo2023gapro} either tend to focus on local structure modeling, bringing dedicated local structure representations like WISGP~\cite{du2023weakly}, or emphasize global relationship modeling, results in a more effective connection between overlapping areas and non-overlapping areas like Gapro~\cite{ngo2023gapro}. 
Both types of modeling are crucial, but there is currently no approach effectively integrating these two aspects. 
Consequently, it is essential to devise a universal network proficient in extracting local structure features efficiently while fostering interactions between overlapping and non-overlapping areas, yielding more precise pseudo-labels.

To achieve the above goals, we introduce a novel pseudo-labeler called Simulation-assisted Transformer (SAFormer), which is trained based on an innovative training strategy called Simulation-assisted Mean Teacher (SMT) and incorporates a special decoder called Local-Global Aware Attention (LGA).
In order to solve the \textbf{first problem} in the previous paragraph, we introduce the SMT. Concretely, student network is directly supervised with definite instance labels for non-overlapping areas, and for overlapping areas, pseudo-labels generated by teacher network are used as supervision. 
As to teacher network, we use EMA to updates its parameters.
This approach yields more accurate predictions for overlapping areas compared to classical statistical methods like GP~\cite{shin2017real}.
Furthermore, to address the challenge of suboptimal pseudo-label quality, we generate simulated overlapping samples using non-overlapping bboxes.
And these simulated samples are used to train a base model, producing weights that serve as the initialization for teacher and student networks. 
This fundamentally equips the network with the ability to distinguish overlapping areas, and the higher quality pseudo-labels generated by the teacher network aid in the rapid training of the Mean Teacher.
Additionally, when applied to multiple datasets, only a brief finetuning is required instead of retraining the pre-trained weights. 
Taking S3DIS~\cite{armeni20163d} as an example, the model's training time decreases from 42 hours to just 1.7 hours.
As to the \textbf{second problem}, we introduce the LGA. Concretely, we first initialize two learnable queries, with each representing one of the two foreground instances. 
We then employ local-structure attention to effectively model the local structure of each instance and aggregate structural relationships within each instance into a holistic representation through queries.
Subsequently, by employing global-context attention, we facilitate the aggregation of global information, especially interactions between the two foreground instances and interactions between overlapping areas and non-overlapping areas. 
Through this design, we can effectively model category, structure, and contextual information adaptively. 
Additionally, we can leverage the response values of the overlapping area points to the queries to remove background points from the overlapping areas.

In summary, the main contributions of this work are as follows: (i) We propose a weakly supervised 3D instance segmentation method called BSNet, which uses bboxes as annotations and devises a novel pseudo-labeler. 
(ii) We design a pioneering pseudo-labeler called SAFormer, which for the first time incorporates the deep neural network and the Mean Teacher paradigm, and innovatively constructs simulated samples to facilitate training. 
Besides, with the help of LGA, SAFormer can accurately predict pseudo-labels for overlapping areas, thus achieving precise weakly supervised 3DIS results.
(iii) Extensive experimental results on two standard benchmarks, ScanNetV2~\cite{dai2017scannet} and S3DIS~\cite{armeni20163d}, verify the superiority of our designs.

\section{Related Work}
In this section, we briefly overview related works on 3D instance segmentation, weakly supervised 3D  instance segmentation and the Mean Teacher paradigm.
\begin{figure*}[!t]
    \centering
    \includegraphics[width=1\linewidth]{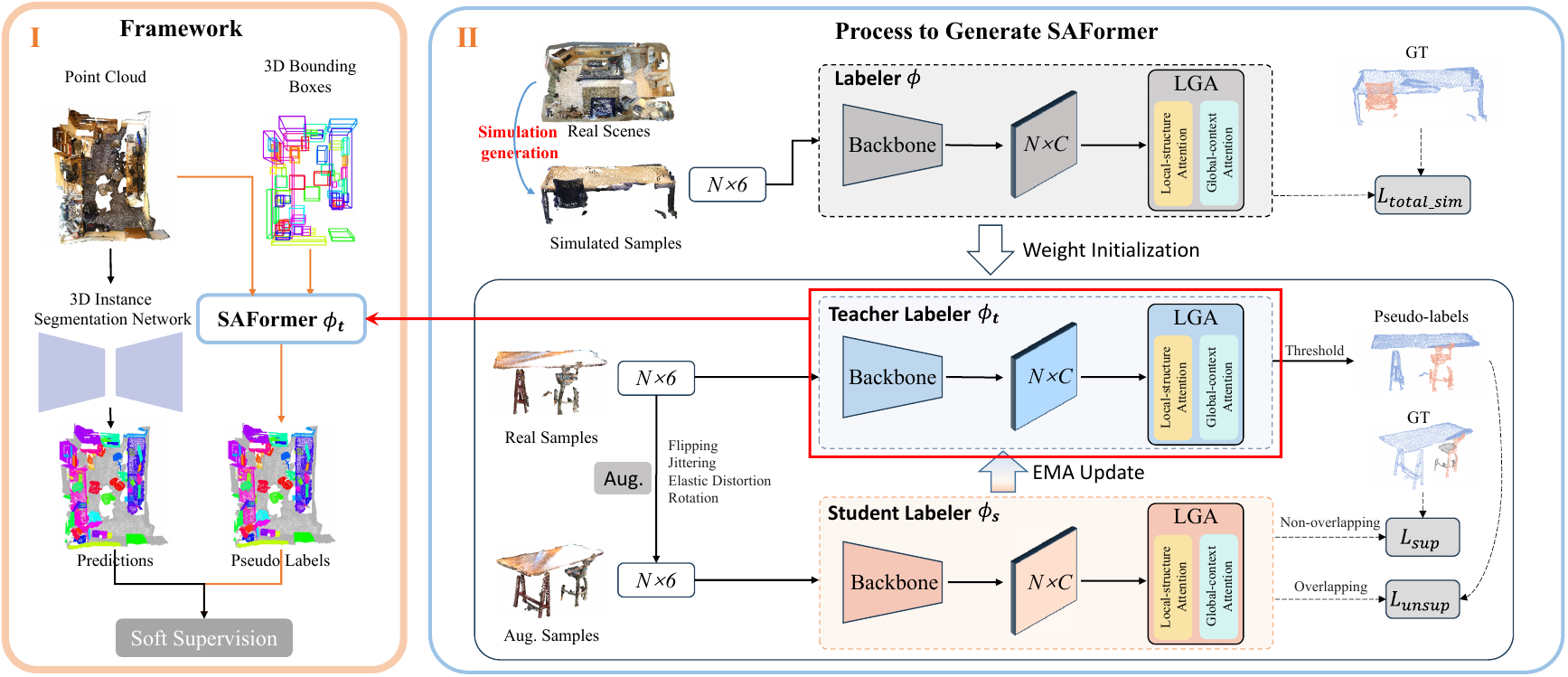}
    \vspace{-2em}
    \caption{
      \textbf{(\Rmnum{1}) The overall framework of our method BSNet. (\Rmnum{2}) The total process to generate an outstanding pseudo-labeler SAFormer.} BSNet is a novel two-step method consisting of generating pseudo instance labels by SPFormer and using the pseudo instance labels to train a 3D instance segmentation network.}
    \label{fig:framework}
    \vspace{-1em}
  \end{figure*}

\noindent\textbf{3D Instance Segmentation.}
3D instance segmentation is a fundamental task for 3D scene understanding, which can be categorized into proposal-based, grouping-based and transformer-based methods. 
Proposal-based methods~\cite{yi2019gspn,yang2019learning,engelmann20203d,liu2020learning} extract 3D bboxes and utilize a mask learning branch to predict the object mask inside each box. 
Grouping-based methods~\cite{jiang2020pointgroup,chen2021hierarchical,liang2021instance,zhong2022maskgroup,vu2022softgroup,wu20223d,ngo2023isbnet} predict semantic categories and geometric offsets for each point, and then employ clustering algorithms to group the points into instances. 
Transformer-based methods~\cite{schult2023mask3d,sun2023superpoint,lu2023query} are the state-of-the-art paradigm, where queries are used to represent instances and global information is aggregated into queries through a transformer decoder~\cite{carion2020end,cheng2021per,cheng2022masked}. 
Although these fully supervised methods have achieved superior performance, they still have significant limitations in practice due to the time-consuming point-wise instance annotation. 
Therefore, some weakly supervised instance segmentation methods have been proposed to alleviate this problem.

\noindent\textbf{Weakly Supervised 3D Instance Segmentation.}
The current weakly supervised 3D instance segmentation methods~\cite{hou2021exploring,xie2020pointcontrast,chibane2022box2mask,du2023weakly,ngo2023gapro} can be divided into two categories: sparse points as annotations and 3D bboxes as annotations. 
Sparse-point annotation methods~\cite{hou2021exploring,xie2020pointcontrast} primarily utilize sparsely labeled point clouds as supervision to train the network. 
In comparison to methods using sparse points as annotations, 3D bboxes provide richer instance information such as category and shape size, enabling the network to better handle instance segmentation tasks. 
Box2mask~\cite{chibane2022box2mask} uses bboxes as supervision, allowing the network to predict bbox for each individual point. 
WISGP~\cite{du2023weakly} leverages 3D local geometric information to generate point-level labels from bbox annotations. 
Gapro~\cite{ngo2023gapro} employs GP~\cite{shin2017real} to model the global similarity relationships between overlapping and non-overlapping regions. 
However, these methods have relatively simple modeling of structure in overlapping samples and do not effectively incorporate category, structure, and contextual information. 
In contrast, our proposed Local-Global Aware Attention enhances the capacity to model both local structures and global relationships.

\noindent\textbf{Mean Teacher Paradigm.}
The Mean Teacher paradigm has been widely researched in various tasks, such as UDA for semantic segmentation~\cite{araslanov2021self,hoyer2022daformer,zhang2021prototypical}, semi-supervised object detection~\cite{xu2021end,mi2022active,wang2023consistent}, weakly supervised object detection~\cite{wang2022omni,wang2023alwod}, UDA for object detection~\cite{cao2023contrastive,li2022cross}, and UDA for person ReID~\cite{ge2020mutual,zheng2021exploiting,han2022delving}. 
This paradigm helps to avoid the interative self-training complicated multi-stage training process. 
SoftTeacher~\cite{xu2021end} introduces the first end-to-end pseudo labeling framework in semi-supervised object detection, gradually improving the quality of pseudo labels during a curriculum. 
To mitigate the issue of low-quality pseudo-labels, CMT~\cite{cao2023contrastive} identifies the alignment and synergy between Mean Teacher and contrastive learning. 
UNRN~\cite{zheng2021exploiting} proposes the estimation and exploitation of the credibility of assigned pseudo-labels for each sample, reducing the impact of noisy pseudo-labels generated by the teacher network. 
Based on the above research, we introduce a Simulation-assisted Mean Teacher approach, which employs the Mean Teacher paradigm to generate stable pseudo-labels in real-time and constructs simulated samples to assist the network in acquiring prior knowledge about overlapping areas. 

\section{Method}
\subsection{Overview}
As illustrated in Figure~\ref{fig:framework}(\Rmnum{1}), the framework of our method begins by generating pseudo object masks for instances in the training set based on bbox annotations. 
Subsequently, these pseudo object masks are employed to train a 3DIS network. 
Throughout the entire process, the most critical step is to generate an outstanding pseudo-labeler to predict pseudo-labels for overlapping areas, as shown in Figure~\ref{fig:framework}(\Rmnum{2}). 
In the generation process, two distinct designs stand out. The first one is the adoption of a unique training strategy called Simulation-assisted Mean Teacher (SMT), which can be divided into two steps: Simulated Sample Generation in Section~\ref{Generation} and Mean Teacher Approach in Section~\ref{SimulationTransfer}. 
The second one is a novel decoder named Local-Global Aware Attention (LGA) in Section~\ref{Local-global}.
First, we generate simulated samples using non-overlapping bboxes from real datasets. These simulated samples are then used to train a labeler $\phi$.
Subsequently, we utilize the weights of $\phi$ to initialize the teacher labeler $\phi_t$ and the student labeler $\phi_s$. 
Finally, we finetune the labelers $\phi_s, \phi_t$ using the Mean Teacher approach to generate pseudo-labels in overlapping areas. The resulting labeler $\phi_t$ is denoted as SAFormer. 
After obtaining the pseudo-labels, we employ them for soft supervision of the 3DIS network.

\subsection{Process to Generate SAFormer}
We develop a novel pseudo-labeler called SAFormer, which accurately predicts labels for overlapping areas, leading to precise results in bbox-supervised 3DIS.
Next, we will sequentially introduce the generation process.

\subsubsection{Simulated Sample Generation}
\label{Generation}
\begin{figure}[!t]
  \centering
  \includegraphics[width=1\linewidth]{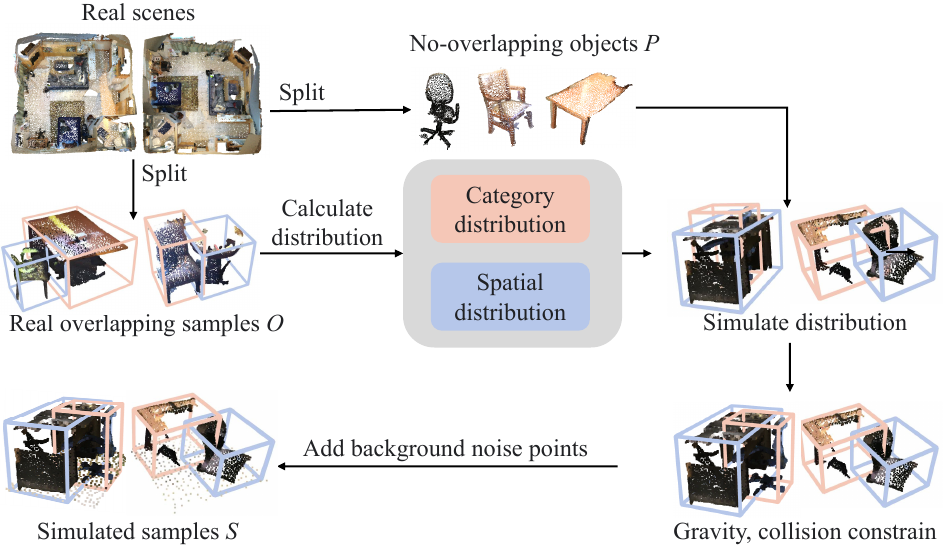}
  \vspace{-2em}
  \caption{
    \textbf{The process of generating simulated samples.}  There are numerous non-overlapping objects ($P$) with definite instance labels in real scenes. We can generate simulated samples ($S$) based on the distribution of real overlapping samples ($O$) and the physical plausibility.}
  \label{fig:generate_pipeline}
  \vspace{-1em}
\end{figure}

The abundant non-overlapping bboxes in ScanNetV2~\cite{dai2017scannet} with definite instance labels allow us to generate simulated overlapping samples.
As illustrated in Figure~\ref{fig:generate_pipeline}, we begin by extracting real overlapping samples $O$ and non-overlapping objects $P$ from real scenes. Subsequently, we conduct an analysis of the class distribution and spatial distribution within these real overlapping samples. To be more specific, we determine which class pairs make up overlapping samples, counting the number $n$ of samples for each class pair and calculate the mean $\mu$  and variance $\sigma$ of the distances between the center points of each class pair.
After obtaining the statistical data, we commence the simulation of the distribution. Firstly, we perform sampling of class pairs based on the distribution of $n$. Assuming the sampled class pair is denoted as $(a, b)$, we then uniformly sample one object point cloud for each of the classes $a$ and $b$ from the set $P$. After obtaining these two point clouds, we perform gaussian sampling based on the corresponding $\mu$ and $\sigma$ to obtain a distance $d$, representing the distance between the two point clouds. Finally, for the sake of simplicity, we directly translate one of the point clouds along the X or Y-axis by the distance $d$.
It is worth noting that, before performing the distance translation, we align the center points of the point cloud pairs.

To better maintain physical plausibility, we make scene adjustments based on the following two principles~\cite{rao2021randomrooms}: 1) gravity: objects should not float in the air; 2) collision: these two objects should not exhibit any collision. Specific details are covered in the supplementary materials. Another purpose of designing the collision constraint is to verify whether two objects can be matched when constructing simulated samples. Concretely, after multiple (with an upper limit of $\mathbf{M}$) distance samplings and collision constraint corrections, some object pairs may still fail to generate overlapping regions. Such pairs are subsequently excluded from use.
Finally, considering that real overlapping samples contain background noise points, we add an appropriate number of $floor$ points to represent the presence of background noise points.

During the aforementioned process, we have obtained simulated overlapping samples $S$.
Currently, we utilize these samples to train a labeler $\phi$.
The labeler $\phi$ primarily consists of two components: a lightweight 3D-UNet based on sparse convolution~\cite{graham2017submanifold,liu2015sparse,graham20183d} and LGA.
In next section, we provide a detailed introduction of LGA.
\subsubsection{Local-Global Aware Attention}
\label{Local-global}
\begin{figure}[!t]
  \centering
  \includegraphics[width=1\linewidth]{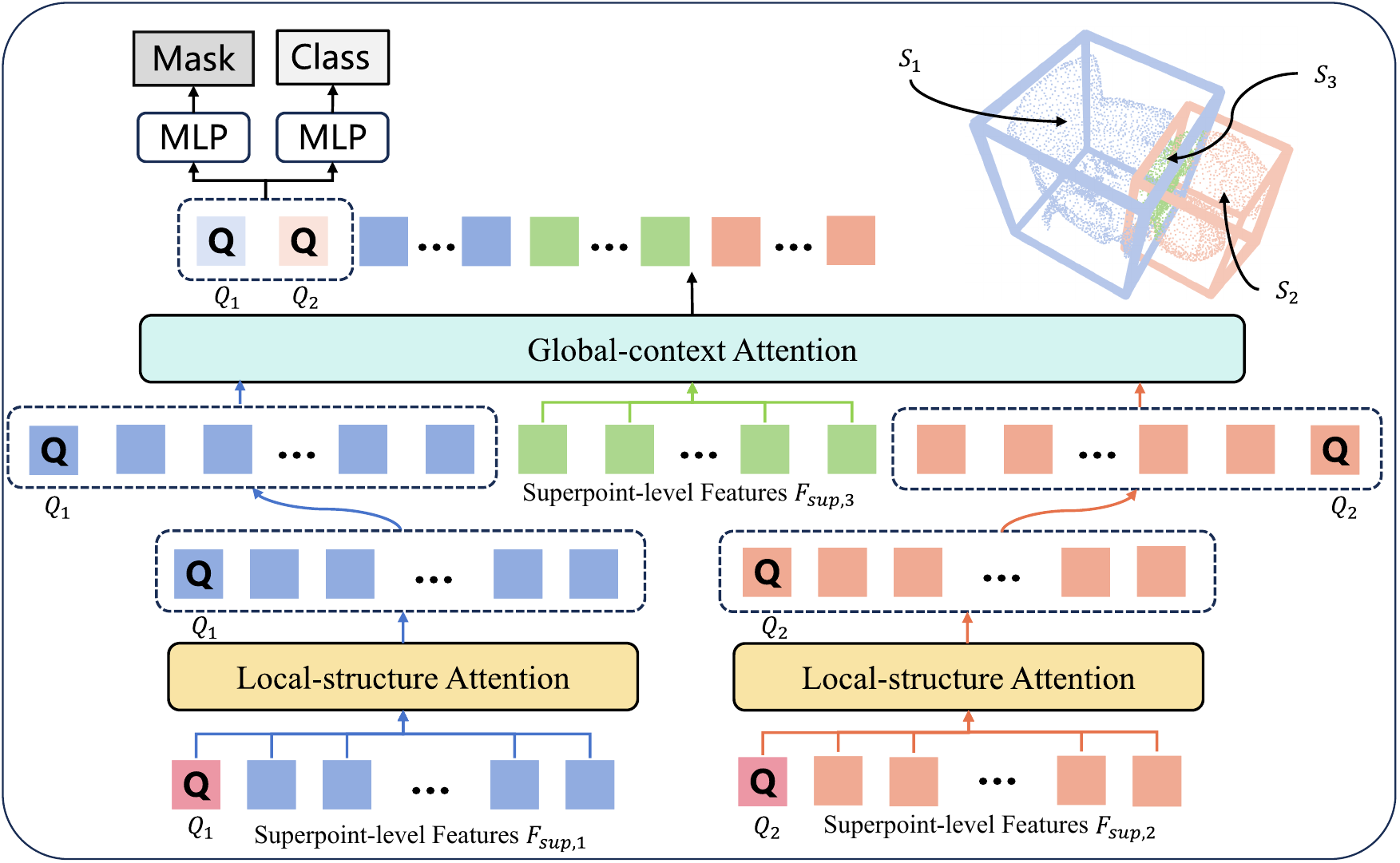}
  \vspace{-2em}
  \caption{
    \textbf{The Local-Global Aware Attention.} Two foreground queries are input into local-structure attention and global-context attention to generate corresponding masks. $S_1, S_2$ represent non-overlapping areas. $S_3$ represents overlapping areas.}
  \label{fig:LGFormer}
  \vspace{-1em}
\end{figure}

As shown in Figure~\ref{fig:LGFormer}, LGA mainly contains local-structure attention and global-context attention.
Assuming that the input point cloud consists of $N$ points, with each point containing position coordinates $(x, y, z)$ and color information $(r, g, b)$.
First, we input the point cloud into a lightweight 3D-UNet to obtain point-level features $F$. 
Subsequently, following SPFormer~\cite{sun2023superpoint}, we aggregate the point-level features $F$ into superpoint-level features $F_{sup}$ using average pooling.
Next, we initialize two learnable queries $Q_1, Q_2$, representing two foreground instances respectively.
To better model local structure, we separately employ the self-attention layer and the feed-forward layer within the non-overlapping areas of different instances. 
This approach ensures that each local region interacts with similar regions belonging to the same instance, significantly enhancing the discriminative and representational capabilities of local structures.
Specifically, we concatenate $F_{sup,1}$ with $Q_1$ and $F_{sup,2}$ with $Q_2$ to form $F_{v,1}, F_{v,2}$, and then input them separately into the self-attention layer, just as follows:
{\small
\begin{equation}
  \label{self-attention}
  F_{v,i}' = {\rm Softmax}(\frac{\mathcal{Q}\mathcal{K}^T}{\sqrt{\mathcal{C}}}) \mathcal{V}, i = 1, 2,
\end{equation}}where $ \mathcal{Q} = F_{v,i}W_q$, $\mathcal{K} = F_{v,i}W_k$, $\mathcal{V} = F_{v,i}W_v$, and $W_q, W_k, W_v$ denote the linear transform matrices for queries, keys, and values, respectively.
Finally we input $F_v'$ into the feed-forward layer to obtain $F_v''$.

After modeling the local structures within each non-overlapping area $S_1, S_2$, and aggregating these structures into a holistic representation through foreground queries, we need to incorporate global information. %
The specific approach involves concatenating the features $F_{v,1}'', F_{v,2}''$, and $F_{sup,3}$. 
Then, we input this concatenated features into the self-attention layer and the feed-forward layer. 
Through this method, we can model relationships between non-overlapping and overlapping areas, between the two foreground instances, and aggregate these global relationships into $Q_1$ and $Q_2$, respectively.
Finally, to classify the overlapping areas, we obtain the masks $M_{ins,1}, M_{ins,2}$ for the two objects by calculating the dot product between $F_{sup,3}$ and $Q_1$, $Q_2$. 
The final mask is obtained through the Sigmoid function followed by a threshold of 0.5:
{\small
\begin{equation}
  \label{sigmoid}
    M_{i} = {\rm Sigmoid}(M_{ins,i}) > 0.5, i = 1, 2.
\end{equation}
}Since $M_1$ and $M_2$ represent two different foreground object masks, for areas where both $M_1$ and $M_2$ are not activated, we classify them as background areas. 
This approach naturally helps the labeler filter out background points, which is an improvement over Gapro~\cite{ngo2023gapro}, as Gapro overlooks the presence of background points.
Furthermore, to better assist the labeler in learning unified knowledge for the same class, we add a class prediction head.

For training on the simulated overlapping samples $S$, since the instance labels are complete, we directly use the shared losses from SPFormer~\cite{sun2023superpoint} and Mask3D~\cite{schult2023mask3d}:
\begin{equation}
  \label{loss_simulated}
   L_{total\_sim} = \lambda_1 L_{cls} + \lambda_2 L_{bce} + \lambda_3 L_{dice},
\end{equation}
where $\lambda_1, \lambda_2, \lambda_3$ are hyperparameters, $L_{cls}$ is the cross-entropy loss, $L_{bce}$ is the binary cross-entropy loss, $L_{dice}$ is the dice loss~\cite{milletari2016v}.
\subsubsection{Mean Teacher Approach}
\label{SimulationTransfer}
During the above process, the labeler $\phi$ has learned prior knowledge about overlapping scenes through training on the simulated samples.
Subsequently, we used the pretrained weights as the initial weights for both the teacher labeler $\phi_t$ and the student labeler $\phi_s$.
Then, we apply data augmentation to the real overlapping samples, including flipping, jittering, elastic distortion, and so on. 
The original samples are fed into the teacher labeler, while the augmented samples are input into the student labeler.
Since the labels for non-overlapping areas in real samples are known, we can directly supervise these areas. 
As for the overlapping areas, to better leverage the predictions of the teacher labeler, we select high-confidence pseudo-labels for the overlapping areas based on a fixed threshold $\tau$.
The teacher labeler updates its parameters using the EMA technique.
With this design, the teacher labeler can continuously update pseudo-labels online and transfer knowledge to the student. 
Simultaneously, the student labeler can employ EMA to transmit the acquired knowledge back to the teacher. 
Going a step further, with the initialization weights obtained through simulation, the teacher labeler gains the ability to distinguish overlapping areas. 
It can generate higher quality pseudo-labels, accelerating the Mean Teacher's training speed. %
Finally, the well-trained teacher labeler $\phi_t$ is referred to as SAFormer, which is used to generate final pseudo-labels for overlapping regions.

For finetuning on the real samples, we adopt a weakly supervised approach. The specific approach can be divided into two parts. First, for non-overlapping areas, where the labels are known but only partial labels of the complete objects, we only supervise the non-overlapping areas.:
{\small
\begin{equation}
  \label{m_no}
    M' = Q_sF_{s,sup,1\cup2}^T,
\end{equation}
}where $Q_s$ represents the instance queries of the student labeler, $1\cup2$ represents the union of $S_1$ and $S_2$,
{\small
\begin{equation}
  \label{loss_real_no}
   L_{sup} = \lambda_2 L_{bce}(M',M_{gt}') + \lambda_3L_{dice}(M',M_{gt}'). 
\end{equation}
}Next, for overlapping areas, we obtain high-confidence pseudo-labels based on a threshold $\tau$ from the teacher labeler and solely supervise the overlapping areas that have corresponding pseudo-labels:
{\small
\begin{gather}
  \label{m_o}
    M_{ps}'' = Q_tF_{t,sup,3}^T \odot (Q_tF_{t,sup,3}^T>\tau), \\
    M'' = Q_sF_{s,sup,3}^T \odot (Q_tF_{t,sup,3}^T>\tau),
\end{gather}
}where $\odot$ represents hadamard product, $Q_t$ represents the instance queries of the teacher labeler, $M_{ps}''$ represents the pseudo-labels of overlapping areas,
{\small
\begin{equation}
  \label{loss_real_o}
  L_{unsup} = \lambda_2L_{bce}(M'',M_{ps}'') + \lambda_3L_{dice}(M'',M_{ps}'').
\end{equation}
}
The total loss for real samples is:
\begin{equation}
  \label{loss_total_real}
  L_{total\_real} = \lambda_1 L_{cls} + L_{sup} + L_{unsup}.
\end{equation}

\subsection{Training a 3DIS Network}
\label{Training}
Due to the fact that the instance labels for points within non-overlapping bboxes and non-overlapping areas of overlapping bboxes are definite,  we can combine these determined instance labels with the pseudo-labels obtained through SAFormer.
Then the combined labels are used to train a 3DIS network.
It's worth noting that since the predicted pseudo-label values $\in[0,1]$, which reflect confidence, employing a soft supervision is a better choice.
Assuming there are $K$ instances, the pseudo masks $M\in[0,1]^{K\times N}$,
{\small
\begin{equation}
  \label{loss_SOFT_BCE}
  L_{bce}' = \frac{\sum_{i = 1}^{K} \sum_{j = 1}^{N}   L_{bce}(M_{pred,i,j},M_{i,j}) *  M_{i,j}}{\sum_{i = 1}^{K} \sum_{j = 1}^{N}M_{i,j}},
\end{equation}
}where $M_{pred}$ represents the results predicted by the 3DIS network.  The total soft loss is:
{\small
\begin{equation}
  \label{loss_soft}
   L_{total\_soft} = \widehat{\lambda_1} L_{cls} + \widehat{\lambda_2}L_{bce}' + \widehat{\lambda_3}L_{dice} + L_{net},
\end{equation}
}where $\widehat{\lambda_1}, \widehat{\lambda_2}, \widehat{\lambda_3}$ are hyperparameters specific to different 3DIS networks, and $L_{net}$ are loss functions unique to different 3DIS networks.

\section{Experiments}
\subsection{Experimental Setup}
\textbf{Datasets and metrics.}
We conduct our experiments on ScanNetV2~\cite{dai2017scannet} and S3DIS~\cite{armeni20163d} datasets.
ScanNetV2 includes 1,613 scenes with 18 instance categories.
Among them, 1,201 scenes are used for training, 312 scenes are used for validation, and 100 scenes are used for test.
S3DIS is a large-scale indoor dataset collected from six different areas, which contains 272 scenes with 13 instance categories.
Following previous works~\cite{vu2022softgroup,ngo2023gapro}, we train on Area 1, 2, 3, 4, 6 and evaluate on Area 5.
AP@25 and AP@50 represent the average precision scores with IoU thresholds 25\% and 50\%,
and mAP represents the average of all the APs with IoU thresholds ranging from 50\% to 95\% with a step size of 5\%.
On ScanNetV2, we report mAP, AP@50 and AP@25. Moreover, we also report the
Box AP@50 and AP@25 results following Gapro~\cite{ngo2023gapro}.
On S3DIS, we report mAP and AP@50.

\textbf{Implementation details.}
\label{Implementation}
The whole method BSNet is trained on a single RTX3090. As to the training setting of the pseudo-labeler SAFormer, first we train 100 epochs on simulated samples with a batch size of 64, which takes about 6 hours.
Next, we fintune 5 epochs on real samples of ScannetV2 training set with a batch size of 64, which takes about 90 minutes.
During inference, it takes approximately 10 minutes to generate pseudo-labels for the entire training set.
As to S3DIS, it takes about 100 minutes for fintuning with 5 epochs and 10 minutes to generate pseudo-labels.
Given that our pseudo-labeler only needs to be trained once on the simulated samples when applied to multiple datasets, so the more datasets we apply, the more efficient the method is.
As to the backbone of SAFormer, we use a lightweight 3D-UNet based on sparse convolution~\cite{graham2017submanifold,liu2015sparse,graham20183d} with 3 blocks and 32 media channels.
At last, we tune the hyperparameters $\mathbf{M}, \tau, \lambda_1, \lambda_2, \lambda_3$ as 8, 0.9, 2, 5, 2.
\subsection{Comparison with state-of-the-art methods}
\begin{figure}[!t]
  \centering
  \vspace{-0.8em}
  \includegraphics[width=1\linewidth]{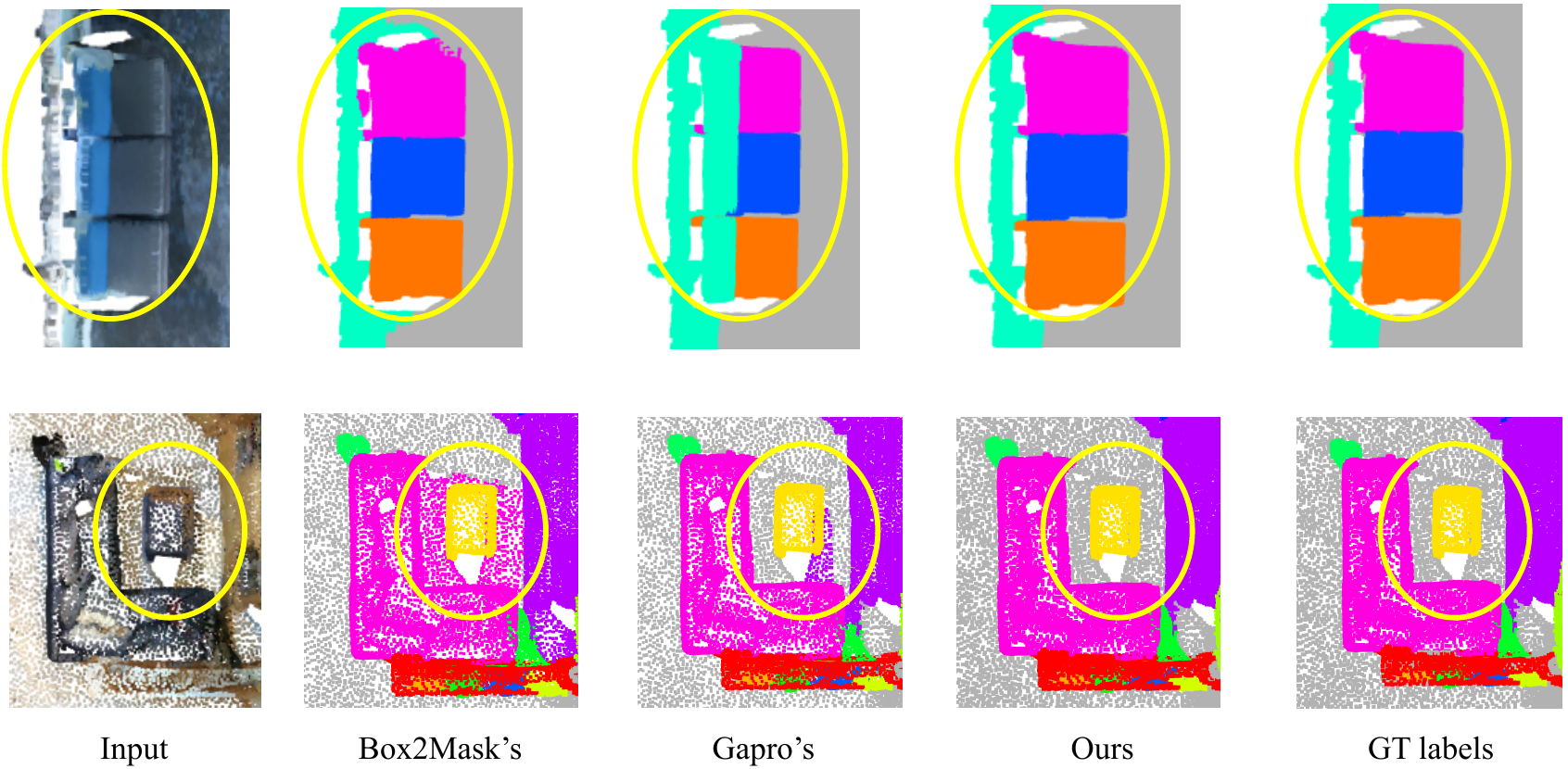}
  \vspace{-2em}
  \caption{
    \textbf{Qualitative results on ScanNetV2 training set. }Our approach produces highly accurate pseudo instance masks, particularly in overlapping areas (yellow circles).}
  \label{fig:visualiztion}
  \vspace{-1em}
\end{figure}

\textbf{ScanNetV2.}
As shown in Table~\ref{table:ScanNetV2}, we compare our approach with existing state-of-the-art methods on the ScanNetV2 validation set.
Attributed to the innovative construction of simulated samples by SMT and the capability of LGA to model local and global information, our proposed SAFormer can generate higher-quality pseudo-labels to supervise the 3DIS network.
Consequently, our box-supervised 3DIS method outperforms other methods by a significant margin in terms of mAP, AP@50 and AP@25.
It is worth emphasizing that our results can achieve 95\% in terms of mAP when compared to the corresponding fully supervised methods. 
This signifies a substantial improvement over previous approaches, which typically achieves only about 90\% performance.
To vividly illustrate the differences between our method and others, we visualize the qualitative results of pseudo-labels in Figure~\ref{fig:visualiztion}. 
From the regions highlighted in yellow circles, it is evident that our method can generate more accurate pseudo-labels for overlapping areas.

\begin{table}[!t]
  \begin{center}
    \footnotesize
    \setlength\tabcolsep{1.6pt}
    \caption{\textbf{Comparison on ScanNetV2 validation set.} \%full indicates the percentage of the current method's performance compared to its corresponding fully supervised method.
    ISBNet$\dagger$ refers that we use the pseudo-labels generated by "Box2Mask~\cite{chibane2022box2mask}: assign points to smaller box" to supervise ISBNet~\cite{ngo2023isbnet}.}
    \label{table:ScanNetV2}
    \vspace{-1.1em}
    \begin{tabular}{ccccccc}
      \toprule
      Method & Sup.  & mAP & \%full & AP@50 & \%full & AP@25  \\
      \midrule
      Mask3D~\cite{schult2023mask3d} & \multirow{5}*{Mask}  & 55.2 & - & 73.7 & - & 83.5  \\
      PointGroup~\cite{jiang2020pointgroup} &  & 34.8 & - & 51.7 & - & 71.3  \\
      SSTNet~\cite{liang2021instance} &   & 49.4 & - & 64.3 & - & 74.0  \\
      ISBNet~\cite{ngo2023isbnet} &   & 54.5 & - & 73.1 & - & 82.5  \\
      SPFormer~\cite{sun2023superpoint} &   & 56.3 & - & 73.9 & - & 82.9  \\
      \midrule
      CSC~\cite{hou2021exploring} & \multirow{2}*{Point}  & 15.9 & 28.8\% & 28.9 & 39.2\% & 49.6  \\
      PointContrast~\cite{xie2020pointcontrast} &   & 27.8 & 50.4\% & 47.1 & 63.9\% & 64.5  \\
      \midrule
      Box2Mask(stand-alone)~\cite{chibane2022box2mask} & \multirow{7}*{Box}  & 39.1 & - & 59.7 & - & 71.8  \\
      ISBNet$\dagger$  & & 41.8 &76.7\%& 64.8&88.6\%&-\\
      WISGP~\cite{du2023weakly} + PointGroup  &   & 31.3 & 89.9\% & 50.2 & 97.1\% & 64.9  \\
      WISGP + SSTNet  &   & 35.2 & 71.3\% & 56.9 & 88.5\% & 70.2  \\
      GaPro~\cite{ngo2023gapro} + ISBNet  &  & 50.6 & 92.8\% & 69.1  & 94.5\% & 79.3  \\
      GaPro + SPFormer  &  & 51.1 & 90.8\% & 70.4  & 95.3\% & 79.9  \\
      Ours + ISBNet  &   & 52.8 & \textbf{96.9\%} & 71.6  & 97.9\% & 82.6  \\
      Ours + SPFormer  &  & \textbf{53.3} & 94.7\% & \textbf{72.7}  & \textbf{98.4\%} & \textbf{83.4}  \\
      \bottomrule
    \end{tabular}
    \vspace{-3em}
  \end{center}
\end{table}

\textbf{S3DIS.}
We evaluate our method on S3DIS using Area 5 in Table~\ref{table:S3DIS}.
Our proposed method achieves superior performance compared to previous methods, with large margins in both mAP and AP@50, demonstrating the effectiveness and generalization of our method.

\textbf{3D object detection results.}
Due to the same level of annotations between box-supervised 3D instance segmentation and fully supervised 3D object detection, our approach can be effectively extended to 3D object detection. 
As illustrated in Table~\ref{table:detection}, our approach performs well in 3D object detection, surpassing the current state-of-the-art methods and the GaPro versions of 3DIS methods. 
It achieves a notable increase of 2.6 in Box AP@50. 

\begin{table}[!t]
  \begin{center}
    \footnotesize
    \setlength\tabcolsep{3pt}
    \caption{\textbf{Comparison on S3DIS on Area 5.} Box2Mask* represents the results of Box2Mask~\cite{chibane2022box2mask} reproduced by Gapro~\cite{ngo2023gapro} on the S3DIS dataset based on their public code.}
    \label{table:S3DIS}
    \vspace{-1.1em}
    \begin{tabular}{cccccc}
      \toprule
      Method & Sup. & mAP & \%full & AP@50 & \%full   \\
      \midrule
      Mask3D~\cite{schult2023mask3d} & \multirow{5}*{Mask} & 56.6 & - & 68.4 & -   \\
      PointGroup~\cite{jiang2020pointgroup} &  & - & - & 57.8 & -  \\
      SSTNet~\cite{liang2021instance} &  & 42.7 & - & 59.3 & -   \\
      SoftGroup~\cite{liang2021instance} &  & 51.6  & - & 66.1 & -  \\
      ISBNet~\cite{ngo2023isbnet} &  & 54.0 & - & 65.8 & -   \\
      \midrule
      Box2Mask*~\cite{chibane2022box2mask} & \multirow{7}*{Box} & 43.6 & - & 54.6 & -   \\
      WISGP~\cite{du2023weakly} + PointGroup  &  & 33.5 & - & 46.8 & 81.0\%   \\
      WISGP + SSTNet  &  &  37.2 & 87.1\% &  51.0 & 86.0\%   \\
      GaPro~\cite{ngo2023gapro} + SoftGroup  &  &  47.0 & 91.1\% & 62.1  & 93.9\%   \\
      GaPro + ISBNet  &  &  50.5 & 93.5\% &  61.2  & 93.0\%   \\
      Ours + SoftGroup  &  & 51.4 & \textbf{99.6\%} & 62.8  & 95.0\%   \\
      Ours + ISBNet  &  & \textbf{53.0} & 98.1\% & \textbf{64.3}  & \textbf{97.7\%} \\
      \bottomrule
    \end{tabular}
    \vspace{-2.8em}
  \end{center}
\end{table}
\subsection{Ablation Study}
The following experiments in Table~\ref{table:designs} are conducted on ISBNet on the validation set of ScanNetV2, while the others are performed on the training set of ScanNetV2. 
\begin{table}[!ht]
  \begin{center}
    \footnotesize
    \setlength\tabcolsep{3pt}
    \vspace{-0.8em}
    \caption{\textbf{3D object detection results on ScanNetV2 validation set. }}
    \label{table:detection}
    \vspace{-1.1em}
    \begin{tabular}{cccc}
      \toprule
      Method & Venue & Box AP@50 & Box AP@25 \\
      \midrule
      VoteNet~\cite{qi2019deep} & ICCV 19  & 33.5 & 58.6  \\
      3DETR~\cite{misra2021end} & ICCV 21 & 47.0 & 65.0  \\
      GroupFree~\cite{liu2021group} & ICCV 21 & 52.8 & 69.1  \\
      HyperDet3D~\cite{zheng2022hyperdet3d} & CVPR 22 & 57.2 & 70.9  \\
      FCAF3D~\cite{rukhovich2022fcaf3d} & ECCV 22 & 57.3 & 71.5  \\
      CAGroup3D~\cite{wang2022cagroup3d}& NeurIPS 22 & 61.3 & 75.1 \\
      \midrule
      GaPro~\cite{ngo2023gapro} + SPFormer~\cite{sun2023superpoint} & ICCV 23  & 65.9  & 78.9   \\
      GaPro + ISBNet~\cite{ngo2023isbnet} & ICCV 23  &  67.0  & 77.1   \\
      Ours + SPFormer & -  & 67.0  & \textbf{80.0}   \\
      Ours + ISBNet  & -  & \textbf{69.6}  & 79.3 \\
      \bottomrule
    \end{tabular}
    \vspace{-2em}
  \end{center}
\end{table}

\textbf{Comparison on pseudo-labels.}
Firstly, we use the metric mAcc to evaluate the quality of pseudo-labels in overlapping areas. 
Assuming that the predicted pseudo-labels of overlapping areas are $P$, the GT of overlapping areas are $P^{gt}$, there are $N$ overlapping areas, there are $M_i$ points in overlapping area $i$, and $\mathbb{I}$ is the indicator function,
{\small
\begin{equation}
  \label{macc}
    {\rm mAcc} = \frac{1}{N} \sum_{i = 1}^{N}\frac{\sum_{j = 1}^{M_i}\mathbb{I} (P_{i,j}==P_{i,j}^{gt})  }{M_{i}}.
\end{equation}
}The higher mAcc represents the better quality of pseudo-labels.
With the help of mAcc, we can explore the impact of different techniques for handling overlapping areas.

\begin{table}[!t]
  \begin{center}
    \footnotesize
    \vspace{-0.8em}
    \caption{\textbf{Quality of pseudo-labels in overlapping areas.} Base refers to utilizing a 3D-UNet and a mask and classification head. LA, GA, MT, SSG represent Local-structure Attention,  Global-context Attention, Mean Teacher, Simulated Sample Generation.}
    \label{table:Quality}
    \vspace{-1.1em}
    \begin{tabular}{lcc}
      \toprule
      Handling of overlapping areas & mAcc  \\
      \midrule
      A: Box2Mask: assign points to smaller box & 24.1\\
      B: Gapro: GP classification with superpoints &38.1 \\
      \midrule
      C0: Base & 41.5 \\
      C1: Ours (LA) & 48.1 \\
      C2: Ours (GA) & 43.5 \\
      C3: Ours (LA + GA) & 52.5 \\
      C4: Ours (LA + GA + MT) & 55.3 \\
      C5: Ours (LA + GA + MT + SSG) & \textbf{59.6} \\
      \bottomrule
    \end{tabular}
    \vspace{-3em}
  \end{center}
\end{table}

As depicted in Table~\ref{table:Quality}, setting B represents the current state-of-the-art technique for handling overlapping areas, and its performance is significantly higher than setting A.
We compare our method SAFormer with these methods and conduct an ablation study of each component in setting C.
In setting C0, attributed to the neural network's strong fitting capability, even without utilizing our proposed SMT and LGA, our base performance still surpasses the current state-of-the-art method Gapro.
In setting C1, we directly train the labeler with a backbone and LA on the real scenes. 
The results show an improvement of 10.0 in mAcc compared to Gapro, indicating that deep neural networks can accurately predict overlapping area labels through dedicated local structure modeling and the accumulation of multiple samples.
In setting C2, we replace LA with GA, resulting in a 5.4 increase in mAcc compared to Gapro.
The results suggest the importance of global information, particularly the interaction between the two foreground instances and between overlapping areas and non-overlapping areas.
In setting C3, we add the design of GA based on C1, resulting in a 4.4 improvement in mAcc.
From the results of C1, C2, and C3, we can conclude that local structure modeling and global relationship modeling complement each other. 
Good local structures form the basis for modeling global relationships, and modeling global relationships can better unleash the potential of good local structures.
In setting C4, to provide stable pseudo-labels for overlapping areas and facilitate information transfer between teacher and student labelers, we add MT. 
The improved performance in mAcc proves its effectiveness.
Finally, to help the labeler gain the ability to distinguish overlapping areas, we add SSG in C5. 
This enables the labeler to predict higher quality pseudo-labels and achieve faster training speed, as shown in Table~\ref{table:SSG2MT}.
\begin{table}[!t]
  \begin{center}
    \footnotesize
    \setlength\tabcolsep{3pt}
    \vspace{-0.8em}
    \caption{\textbf{Effect of our method's components.} Our pseudo-labels: the pseudo-labels generated by our proposed pseudo-labeler SAFormer. Soft loss: the soft loss proposed in Section~\ref{Training}.}
    \label{table:designs}
    \vspace{-1.1em}
    \begin{tabular}{cc|ccc}
      \toprule
      Our pseudo-labels& Soft loss & mAP & AP@50 & AP@25  \\
      \midrule
      \ding{55} & \ding{55} &38.1& 59.1& 72.7 \\ 
      \ding{51}&\ding{55} &52.3& 71.2& 82.1 \\
      \ding{51}& \ding{51} &\textbf{52.8}& \textbf{71.6}& \textbf{82.6} \\
      \bottomrule
    \end{tabular}
    \vspace{-2em}
  \end{center}
\end{table}

\begin{table}[!t]
  \begin{center}
    \footnotesize
    \vspace{-0.8em}
    \caption{\textbf{Effect of different pseudo-label utilization methods.} 
    Base refers that pseudo-labels are directly used to train the 3DIS network. Iterative self-training refers that updating pseudo-labels offline after each training round and then using the updated pseudo-labels to further optimize the labeler. After multiple iterations, the latest pseudo-labels are used to train the 3DIS network.}
    \label{table:utilization}
    \vspace{-1.1em}
    \begin{tabular}{c|c}
      \toprule
      Method& mAcc \\
      \midrule
      Base & 52.5 \\
      Iterative self-training& 52.6 \\ 
      Mean Teacher& \textbf{55.3} \\     
      \bottomrule
    \end{tabular}
    \vspace{-1em}
  \end{center}
\end{table}
\begin{table}[!t]
  \begin{center}
    \footnotesize
    \vspace{-0.8em}
    \caption{\textbf{Effect of different steps in SSG.} SD, GCC, ABP represent simulating distribution, gravity-collision constrain, adding background points respectively. 
    }
    \label{table:stepssimulated}
    \vspace{-1.1em}
    \begin{tabular}{ccc|c}
      \toprule
      SD&GCC& ABP &mAcc \\
      \midrule
      \ding{51} & \ding{55} & \ding{55} & 58.5\\
      \ding{51} &\ding{51} &\ding{55}  & 59.3\\
      \ding{51} &\ding{51} &\ding{51} & \textbf{59.6}\\
      \bottomrule
    \end{tabular}
    \vspace{-1em}
  \end{center}
\end{table}

\textbf{Effect of our method's components.} Table~\ref{table:designs} shows 3DIS results with different components.
In the first row, we evaluate the approach of ignoring overlapping areas during training and only using the determined regions as pseudo-labels.
The second row showcases the efficacy of the pseudo-labels produced by our proposed labeler SAFormer, resulting in a 14.2 improvement in mAP.
In the last row, to validate the impact of the soft loss, we conduct a corresponding ablation experiment and achieve a performance boost of 0.5 in mAP.
\begin{figure}[!t]
  \centering
  \vspace{-0.8em}
  \includegraphics[width=1\linewidth]{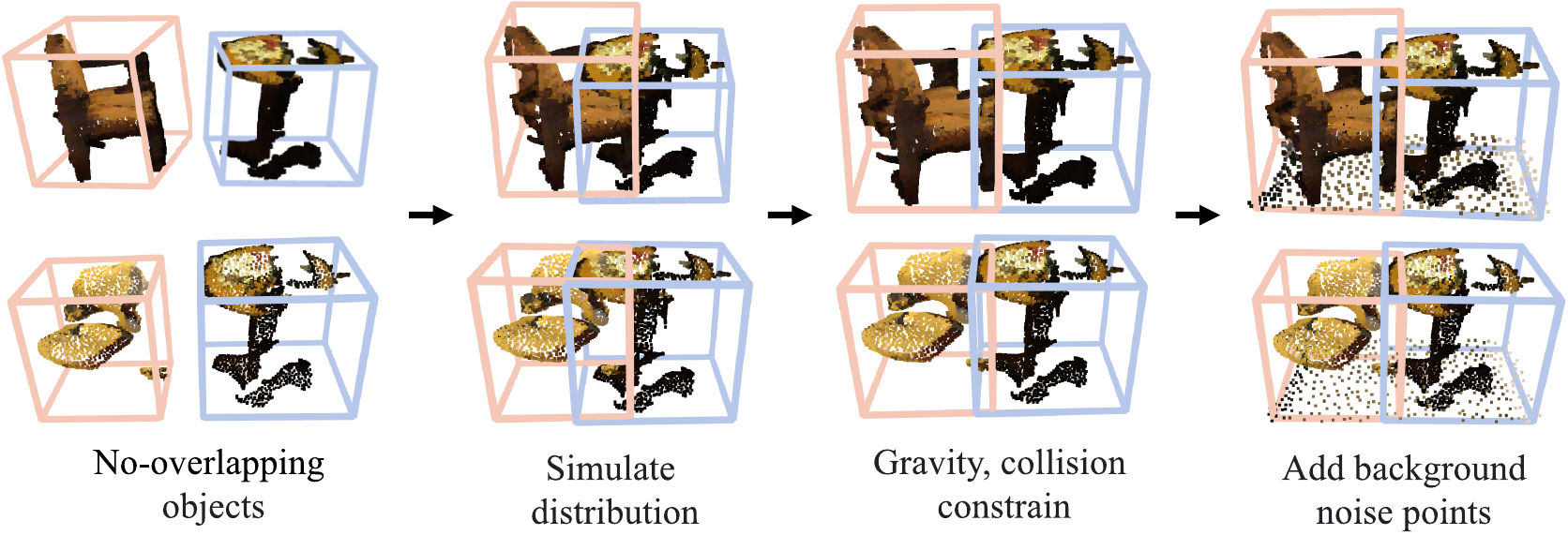}
  \vspace{-2em}
  \caption{
    \textbf{Qualitative visualization results of our SSG. }}
  \label{fig:visualiztionsimulated}
  \vspace{-1.5em}
\end{figure}

\textbf{Effect of different pseudo-label utilization methods. } 
As shown in Table~\ref{table:utilization}, we observe that iterative self-training contributes minimally to performance improvement, whereas Mean Teacher results in a 2.8 increase in mAcc.
The findings highlight that Mean Teacher can generate higher quality pseudo-labels by facilitating information transfer between the student and teacher labeler.

\textbf{Effect of different steps in SSG. } 
Table~\ref{table:stepssimulated} illustrates that as the simulated overlapping samples become more realistic, the quality of pseudo-labels is getting better. 
It's worth noting that adding background points results in a 0.3 increase in mAcc. 
This is partly because it makes the samples more realistic. 
On the other hand, it is because our designed mask activation using sigmoid function can naturally filter out background points.
In order to illustrate the generation process more vividly, we visualize the qualitative results in Figure~\ref{fig:visualiztionsimulated}.
It is shown that the generated simulated samples successfully combine the individual 3D shapes in a meaningful way.

\textbf{Effect of SSG. }
As shown in Table~\ref{table:SSG2MT}, with the assistance of SSG, the labeler can predict higher quality pseudo-labels. 
Moreover, owing to the labeler's initialization with simulated samples, the teacher labeler can furnish more stable and accurate pseudo-labels in the early stages of training, thereby expediting the overall training process.

\textbf{Effect of a class head. }
Based on Table~\ref{table:class}, it can be deduced that the incorporation of a class head helps the labeler acquire unified representations for the same class, resulting in more precise pseudo-labels.
\begin{table}[!t]
  \begin{minipage}{0.48\textwidth}
    \vspace{-0.8em}
    \begin{minipage}[c]{0.48\textwidth}
      \centering
      \captionof{table}{\textbf{Effect of SSG to MT.}}
      \vspace{-1.1em}
      \label{table:SSG2MT}
      \setlength\tabcolsep{3pt}
      \scalebox{0.8}{\begin{tabular}{c|cc}
        \toprule
        Setting & Training time(h) & mAcc   \\
        \midrule
        w/o SSG & 40 & 55.3 \\
        w SSG  & \textbf{1.5} & \textbf{59.6}\\
        \bottomrule
      \end{tabular}}
      \end{minipage}
      \begin{minipage}[c]{0.48\textwidth}
      \centering
      \captionof{table}{\textbf{Effect of a class head.}}
      \label{table:class}
      \vspace{-1.1em}
      \scalebox{0.8}{\begin{tabular}{c|c}
        \toprule
        Setting  & mAcc  \\
        \midrule
        w/o class head & 59.2\\
        w class head  & \textbf{59.6}\\
        \bottomrule
      \end{tabular}}
    \end{minipage}
    \end{minipage}
    \vspace{-1em}
\end{table}
\begin{table}[!t]
  \begin{center}
    \footnotesize
    \caption{\textbf{Comparison of parameters and training time.} $\rm{T}$ represents the total training time, which includes the time to generate pseudo-labels and the time to train 3DIS network with the pseudo-labels. 
    $\rm{\widehat{P}}$ represents the pseudo-labeler parameters, $\rm{P}$ represents the corresponding 3DIS network parameters, and \%full denotes the proportion of $\rm{\widehat{P}}$ to $\rm{P}$. }
    \label{table:trainingtime}
    \vspace{-1.1em}
    \begin{tabular}{c|cccc}
      \toprule
      Method & $\rm{T}$(h) & $\rm{\widehat{P}}$(M) & $\rm{P}$(M)& \%full  \\
      \midrule
      Gapro + ISBNet & 150 &- & 30.7& - \\
      Gapro + SPFormer & 80 & -&17.6 &- \\
      Ours + ISBNet & 72 &2.4 &30.7& 7.8\%\\
      Ours + SPFormer & 37& 2.4& 17.6&13.6\% \\
      \bottomrule
    \end{tabular}
    \vspace{-2.8em}
  \end{center}
\end{table}
\subsection{Parameters and Training Time Analysis}
Table~\ref{table:trainingtime} reports the parameters and the training time on ScanNetV2 training set.
For a fair comparison, the reported training time is measured on the same device.
Our pseudo-labeler utilizes only about 10\% of the corresponding 3DIS network parameters, making it very lightweight. And in terms of time, it is less than half of Gapro's. 
This can be attributed to different self-training ways and different objects to which self-training is applied.
Gapro performs iterative self-training on pseudo-labeler and 3DIS network, while our method performs Mean Teacher self-training only on pseudo-labeler.
Therefore, our method not only eliminates the high time cost caused by repeated training of the 3DIS network, but also greatly alleviates the training time of Mean Teacher through the design of SMT.

\section{Conclusion}
In this paper, we propose the Box-Supervised Simulation-assisted Mean Teacher for 3D Instance Segmentation, which devises a novel pseudo-labeler called SAFormer.
To the best of our knowledge, SAFormer is the first labeler incorporating the deep neural network and Mean Teacher in this task, and innovatively constructs simulated samples to facilitate training. 
Furthermore, the well-designed transformer decoder LGA effectively models local structures and global relationships of point clouds.
Extensive experiments conducted on two widely used box-supervised 3D instance segmentation benchmarks demonstrate the superior performance of our method.

\section{Acknowledgements}
This work was partially supported by the Youth Innovation Promotion Association CAS 2018166, Anhui Provincial Natural Science Foundation (Grant 2308085QF222) and National Defense Basic Scientific Research program (JCKY2021601B013).

{
    \small
    \bibliographystyle{ieeenat_fullname}
    \bibliography{egbib}
}


\end{document}